\documentclass[letterpaper,10pt,conference]{ieeeconf}  

\IEEEoverridecommandlockouts                              

\usepackage{amsmath} 
\usepackage{graphicx}
\usepackage{amssymb}
\usepackage[utf8]{inputenc}
\usepackage{cite}
\usepackage{hyperref}

\title{\LARGE \bf Symmetry-Aware Fusion of Vision and Tactile Sensing via \\ Bilateral Force Priors for Robotic Manipulation}

\author{
Wonju Lee$^{1,\dagger}$, Matteo Grimaldi$^{1}$, and Tao Yu$^{2}$%
\thanks{$^{\dagger}$ denotes the corresponding author}%
\thanks{$^{1}$DexAI, Emergent Business Unit, Analog Devices Inc., Limerick, Ireland
({\tt\small wonju.lee@analog.com, matteo.grimaldi@analog.com})}%
\thanks{$^{2}$DexAI, Emergent Business Unit, Analog Devices Inc., Boston, MA, USA
({\tt\small tao.yu@analog.com})}%
}

\begin{document}

\maketitle
\thispagestyle{empty}
\pagestyle{empty}


\begin{abstract}
Insertion tasks in robotic manipulation demand precise, contact-rich interactions that vision alone cannot resolve. While tactile feedback is intuitively valuable, existing studies have shown that naïve visuo-tactile fusion often fails to deliver consistent improvements. In this work, we propose a Cross-Modal Transformer (CMT) for visuo-tactile fusion that integrates wrist-camera observations with tactile signals through structured self- and cross-attention. To stabilize tactile embeddings, we further introduce a physics-informed regularization that encourages bilateral force balance, reflecting principles of human motor control. Experiments on the TacSL benchmark show that CMT with symmetry regularization achieves a $96.59\%$ insertion success rate, surpassing naïve and gated fusion baselines and closely matching the privileged ``wrist + contact force'' configuration ($96.09\%$). These results highlight two central insights: (i) tactile sensing is indispensable for precise alignment, and (ii) principled multimodal fusion, further strengthened by physics-informed regularization, unlocks complementary strengths of vision and touch, approaching privileged performance under realistic sensing.
\end{abstract}

\section{Introduction}\label{sec:intro}
Robotic insertion is a long-standing benchmark in contact-rich manipulation, requiring accurate perception and fine-grained control under uncertainty. Vision-based policies, enabled by deep architectures such as CNNs~\cite{he2016deep} and Vision Transformers~\cite{dosovitskiy2021image}, excel at global scene understanding and object localization. However, in insertion tasks they often fail to capture subtle physical interactions---such as micro-slippage, compliance, or misalignment during contact---and remain sensitive to occlusion, lighting, and incomplete geometry~\cite{huang2021multi,inoue2017assembly,lee2020multimodal}.

Tactile sensing complements vision by directly measuring local contact states. GelSight~\cite{yuan2017gelsight}, DIGIT~\cite{donlon2020digit}, and TacTip~\cite{ward2021tactile} encode surface deformations that can be transformed into geometric and force-distribution descriptors. Empirical studies confirm their value in grasp stability~\cite{calandra2017feeling}, slip detection~\cite{li2020learning}, and dexterous manipulation~\cite{li2022learning}. For insertion, tactile feedback is particularly crucial: contact forces signal misalignments and socket interactions that vision alone cannot infer.

\begin{figure*}[t]
\begin{center}
\includegraphics[width=5.2in]{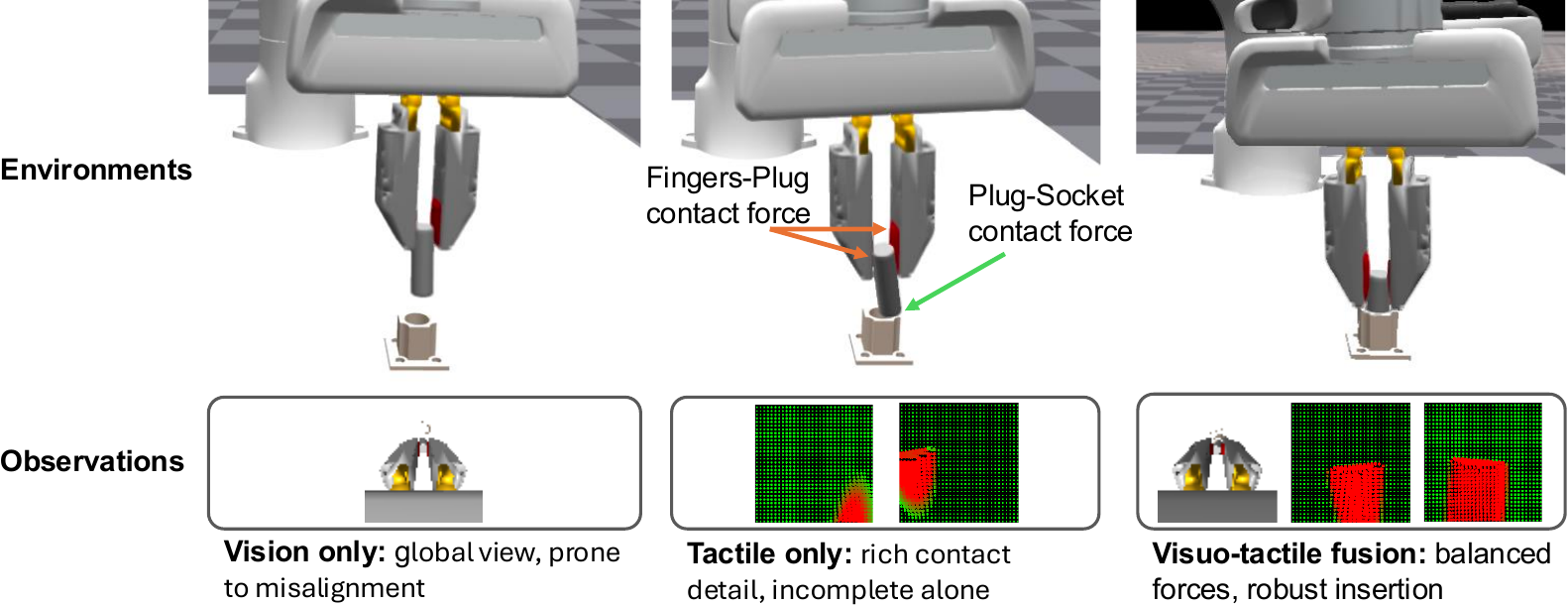}
\end{center}
\vspace{-5mm}
\caption{Comparison of observation modalities for robotic insertion policies. \textbf{Left}: Vision-only input provides global alignment cues but lacks local precision. \textbf{Center}: Tactile-only input encodes fine-grained force signals critical for corrective actions. \textbf{Right}: Visuo-tactile fusion integrates coarse visual guidance with detailed tactile feedback, achieving robust insertion by exploiting complementary strengths.}
\label{fig:motiv}
\vspace{-4mm}
\end{figure*}

Fig.~\ref{fig:motiv} illustrates the complementary roles of different observation modalities. Vision (left) provides coarse global alignment but misses fine corrections. Tactile sensing (center) captures local force patterns essential for precise adjustments. Visuo-tactile fusion (right) combines these strengths, yielding robust insertion behavior. Quantitatively, augmenting either low-dimensional state inputs or wrist-camera observations with tactile signals improves insertion success by $+2.2\%$ and $+2.8\%$, respectively (Table~\ref{tb:perf}), underscoring the indispensability of contact feedback.
We further observe in extended experiments on the screw task (Appendix~\ref{sec:appendix2}) that tactile sensing alone achieves perfect success, reinforcing its role as a primary modality in contact-rich manipulation.

Yet, effectively leveraging tactile signals is challenging. TacSL~\cite{tacsl2024} reported negligible or even negative gains with direct feature concatenation, highlighting two issues: (i) the difficulty of synchronizing heterogeneous representations, and (ii) the risk of diluting modality-specific cues. These limitations motivate the need for structured fusion strategies that respect the role of each modality and exploit their complementarity in a task-aware manner.

Recent works have investigated attention mechanisms for this purpose: self-attention for intra-modal correspondences~\cite{chen2022vtfsa}, cross-modal alignment~\cite{wu2025convitac,zhang2025vitacformer,liu2025adaptivefusion}, timing-aware fusion~\cite{an2024interplay}, and force-guided weighting~\cite{poppi2021adaptac,li2025adaptivefusion}. Together, these advances suggest that principled attention-based fusion, rather than naïve concatenation, is key to exploiting visuo-tactile synergy.

In this paper, we propose a {\it symmetry-aware, physics-informed visuo-tactile fusion framework} tailored for robotic insertion. Our design is motivated by human motor control, where bilateral force balance ensures stability~\cite{morasso1981arm, bizzi1991modular}. Inspired by these principles, we introduce a \textbf{bilateral force regularization} that explicitly enforces vertical symmetry between left and right finger forces. This physics-informed inductive bias stabilizes grasps and reduces lateral misalignments during insertion. 

The symmetry-regularized tactile embeddings are fused with visual features via a Cross-Modal Transformer (CMT), which applies hierarchical self- and cross-attention to integrate global visual context with local, physically consistent tactile feedback. By combining structured attention with physics-informed regularization, our framework yields smoother, more robust insertion trajectories compared to vision-only, naïve fusion, and TacSL baselines.

Our contributions are threefold:
\begin{itemize}
\item \textbf{Methodological novelty}: We propose a Cross-Modal Transformer (CMT) that integrates vision and tactile cues via hierarchical self- and cross-attention, addressing synchronization challenges that hinder naïve fusion.
\item \textbf{Physics-informed regularization}: We introduce a bilateral force-symmetry constraint inspired by human motor control~\cite{morasso1981arm,bizzi1991modular}, which encodes a physics-based inductive bias for balanced grasping and stable insertion.
\item \textbf{Experimental validation}: We demonstrate superior insertion performance over naïve and gated fusion, achieving $96.59\%$, nearly matching the privileged wrist+force configuration ($96.09\%$).
\end{itemize}

\section{Related Work}\label{sec:related}

\subsection{Vision-based Manipulation}
Vision has long been the primary sensing modality for robotic manipulation due to its scalability and the maturity of visual deep learning. CNN- and Transformer-based policies~\cite{he2016deep,dosovitskiy2021image} achieve strong results in recognition and pick-and-place. However, vision-only pipelines often struggle in insertion tasks due to depth ambiguity, occlusion, and the absence of physical cues such as contact forces~\cite{huang2021multi,inoue2017assembly,lee2020multimodal}. Large-scale visuomotor benchmarks such as RoboNet~\cite{dasari2019robonet} and RLBench~\cite{james2020rlbench} reinforce this limitation, motivating the integration of tactile feedback in contact-rich domains.

\subsection{Tactile Sensing for Contact-rich Tasks}
Tactile sensing directly captures local interaction states, including contact geometry, slip, and force distributions. Vision-based tactile sensors such as GelSight~\cite{yuan2017gelsight}, DIGIT~\cite{donlon2020digit}, and TacTip~\cite{ward2021tactile} provide high-resolution deformation maps that CNN backbones can transform into compact descriptors. These signals have been shown to improve grasp stability~\cite{calandra2017feeling,mahler2019learning}, slip detection~\cite{li2020learning}, and dexterous in-hand manipulation~\cite{li2022learning}. 
Beyond optical sensors, force–torque sensors~\cite{xu2019improving} and capacitive arrays~\cite{yamaguchi2017implementing} have been applied to insertion and assembly, though they offer lower spatial resolution. High-frequency tactile datasets further highlight the importance of transient slip and compliance cues~\cite{she2020intact}. For insertion in particular, augmenting visual or low-dimensional policies with tactile input consistently improves success, underscoring the indispensability of tactile sensing for alignment and frictional dynamics (Table~\ref{tb:perf}).

\subsection{Visuo-Tactile Fusion Strategies}
The central challenge in visuo-tactile fusion lies in reconciling heterogeneous feature spaces while preserving modality-specific information. Naïve concatenation, as reported in TacSL~\cite{tacsl2024}, often yields negligible or negative gains. To address this, structured fusion approaches have emerged. VTFSA~\cite{chen2022vtfsa} applies self-attention across visuo-tactile features, while ConViTac~\cite{wu2025convitac} and ViTacFormer~\cite{zhang2025vitacformer} employ cross-modal Transformers for alignment. Adaptive methods such as AdapTac~\cite{poppi2021adaptac} and its extensions~\cite{li2025adaptivefusion} use predictive force cues to reweight modalities dynamically. More recently, VTLA~\cite{vtla2025} and OmniVTLA~\cite{omni_vtla2025} integrate language to guide fusion and improve task generalization. Despite these advances, most existing methods treat tactile signals as features to be adaptively weighted, rather than embedding explicit physical principles into the learning process.

\subsection{Symmetry and Physical Priors}
Human motor control studies highlight the role of bilateral symmetry and modular control in achieving stable and coordinated movements. Morasso~\cite{morasso1981arm} showed that reaching movements are spatially organized, while Bizzi et al.~\cite{bizzi1991modular} demonstrated modular synergies in spinal control. 
In robotics, Ilonen et al.~\cite{ilonen2014symmetry} exploited symmetry priors for tactile object reconstruction, and Su et al.~\cite{kwiatkowski2019symmetry} applied symmetry constraints for dual-arm manipulation. More broadly, physics-informed machine learning~\cite{karniadakis2021physics} has emerged as a paradigm for embedding physical constraints directly into neural architectures. 
Our work contributes to this line by introducing bilateral force symmetry as a physics-informed regularizer for visuo-tactile policy learning, rather than for perception or state estimation. This transforms tactile signals into structured, physically meaningful representations that stabilize grasps and reduce jamming during insertion.

\begin{figure*}[t]
  \begin{center}
    \includegraphics[width=6.2in]{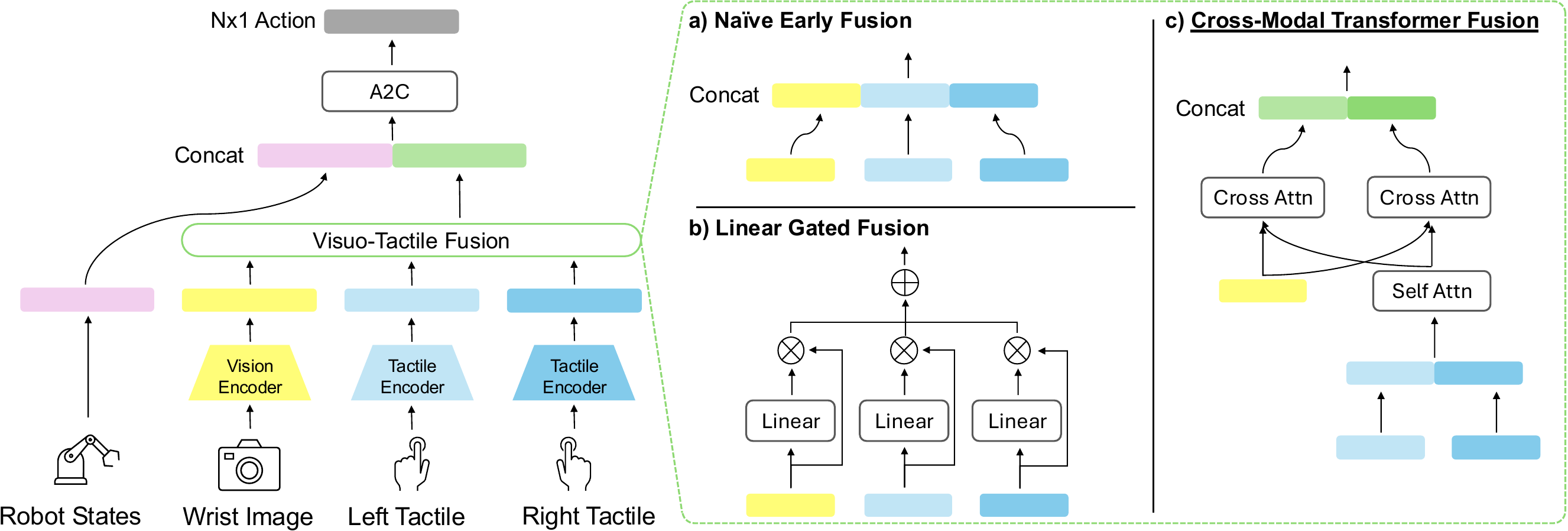}
  \end{center}
  \vspace{-2mm}
    \caption{Overview of visuo-tactile fusion architectures. (a) Naïve concatenation of embeddings, which risks diluting modality-specific signals. (b) Gated fusion with linear layers that adaptively weight neuronal contributions. (c) The proposed Cross-Modal Transformer (CMT), which embeds symmetry-aware tactile encoding and integrates vision and touch via cross-attention.}
  \label{fig:method}
  \vspace{-4mm}
\end{figure*}

\subsection{Positioning of Our Work}
Prior work has (i) highlighted the limitations of vision-only policies, (ii) shown the utility of tactile sensing, and (iii) explored structured visuo-tactile fusion strategies. However, existing approaches either assume symmetry for state estimation and 3D reconstruction~\cite{ilonen2014symmetry} or employ force-guided attention to reweight modalities adaptively~\cite{poppi2021adaptac,li2025adaptivefusion}. 
These methods focus on object modeling or predictive force tasks rather than direct policy learning. In contrast, our framework leverages bilateral symmetry as a physics-informed regularization that stabilizes tactile embeddings and integrates them with vision through hierarchical attention. This combination yields a principled, policy-driven approach that advances beyond naïve concatenation and prior adaptive fusion mechanisms, enabling robust performance in contact-rich insertion.

\section{Proposed Method: Symmetry-Aware Visuo-Tactile Fusion} \label{sec:prop}
We present a novel visuo-tactile fusion framework for robotic insertion that combines global alignment cues from vision with local corrective feedback from tactile sensing. 
Our method introduces a \textbf{physics-informed symmetry prior} to regularize tactile embeddings, which encourages balanced bilateral forces and mitigates jamming during insertion. 
This prior is integrated into a \textbf{Cross-Modal Transformer (CMT)} architecture that employs hierarchical self- and cross-attention, ensuring structured fusion between modalities. 
The overall framework is summarized in Fig.~\ref{fig:method}.

\subsection{Problem Formulation}
We formulate insertion as a partially observable Markov decision process (POMDP) defined by $(S,A,O,T,R)$, where $S$ is the latent state, $A$ continuous robot actions, $O$ multimodal observations, $T$ transition dynamics, and $R$ the sparse reward function.
The observation $O$ comprises (i) RGB wrist camera inputs for global alignment and (ii) tactile force fields from the gripper fingers for local contact sensing.
The policy $\pi_\theta(a|o)$ aims to maximize the expected discounted return as
\begin{equation}
J(\theta) = \mathbb{E}_{\pi_\theta}\Bigg[\sum_{t=0}^{t_{\max}} \gamma^t r_t \Bigg],
\end{equation}
with $\gamma \in (0,1)$ and reward $r_t=1$ only upon successful insertion.

\subsection{Residual Tactile Encoding with Symmetry Priors}
Let the raw tactile forces at time $t$ be $\bar{f}_t^L,\bar{f}_t^R \in \mathbb{R}^d$.  
To account for possible object asymmetry, we define residual forces relative to calibrated reference signals by
\begin{equation}
f_t^L = \bar{f}_t^L - f_{ref}^L, \quad f_t^R = \bar{f}_t^R - f_{ref}^R,
\end{equation}
where $f_{ref}^L, f_{ref}^R$ are prior forces obtained either through a short calibration contact or set to zero for symmetric objects.  
This formulation generalizes symmetry-aware balancing: symmetric objects correspond to $f_{ref}^L=f_{ref}^R=0$, while asymmetric objects can be handled by anchoring to calibrated reference distributions.  

The calibration is performed in a pre-insertion step by gently grasping the object to record the stable, pre-contact force distribution. This measured force profile serves as a physics-informed baseline, allowing the policy to learn corrective actions based on deviations from this nominal state rather than on the absolute force values themselves. While our current evaluation focuses on symmetric objects, this residual formulation is designed to be robust to object asymmetry and variations in grasp pose, generalizing the core principle of force balancing to a wider class of manipulation tasks.

The residuals are encoded via backbones into embeddings $h_t^L, h_t^R$.  
To capture bilateral consistency, we perform self-attention over concatenated tactile features as
\begin{equation}
z_t^T = \text{Attn}(W_q[h_t^L;h_t^R],\, W_k[h_t^L;h_t^R],\, W_v[h_t^L;h_t^R]),
\end{equation}
yielding a residual-aware tactile representation that normalizes for asymmetry before fusion. This step enforces coherent intra-modal structure before cross-modal fusion.

\subsection{Visuo-Tactile Cross-Attention}
Camera observations $v_t$ are embedded as $z_t^V=\phi^V(v_t)\in\mathbb{R}^k$. 
We then apply cross-attention with vision as \textit{query} (guiding alignment) and tactile as \textit{key/value} (providing corrective feedback) as
\begin{equation}
z_t^{VT} = \text{Attn}(W_q^V z_t^V,\, W_k^T z_t^T,\, W_v^T z_t^T).
\end{equation}
This asymmetry reflects our design choice: vision provides global context, while tactile sensing supplies fine-grained local corrections. 
By stacking intra-tactile self-attention and visuo-tactile cross-attention, the CMT achieves \textbf{hierarchical fusion}, structurally modeling the roles of each modality.

\label{sec:prop}
\begin{figure}[t]
  \begin{center}
    \includegraphics[width=3.2in]{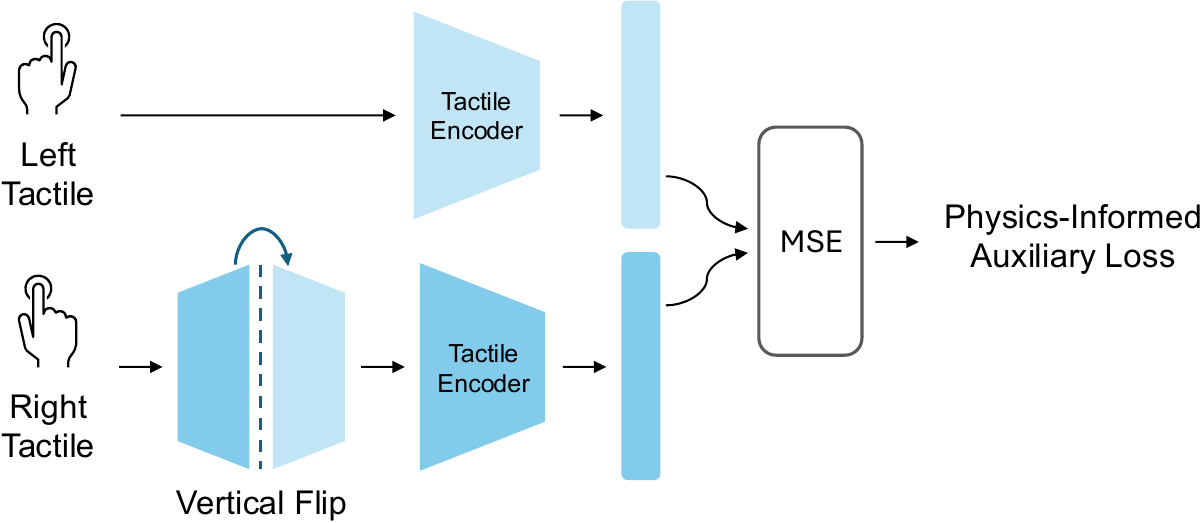}
  \end{center}
  \vspace{-4mm}
  \caption{Physics-informed symmetry regularization. The right tactile map is vertically flipped and encoded as $\tilde{h}_t^R$, then compared with $h_t^L$. The mean squared error loss penalizes deviations, encouraging bilateral consistency. This auxiliary objective stabilizes grasp forces before insertion and reduces lateral misalignment during insertion.}
  \label{fig:sym_loss}
  \vspace{-4mm}
\end{figure}

\begin{table*}[t]
\caption{Success rates under privileged and reduced settings across different sensory modalities and fusion strategies. Improvements over the reduced baseline are shown in parentheses.}
\vspace{-1mm}
\centering
\begin{tabular}{lcccccc}
\hline
Method                 & Privileged & Reduced & Contact forces & Wrist & Tactile & Success rate (\%) \\ \hline \hline
Privileged               & \checkmark          &         &                &       &         & 96.74 $\pm$ 1.63            \\
+ Contact forces            & \checkmark          &         & \checkmark              &       &         & 98.96 $\pm$ 0.83 (+2.22)     \\ \hline \hline
Tactile                &            & \checkmark       &                &       & \checkmark       & 91.41 $\pm$ 5.51            \\
Wrist                  &            & \checkmark       &                & \checkmark     &         & 93.23 $\pm$ 2.00            \\
Wrist + Contact forces &            & \checkmark       & \checkmark              & \checkmark     &         & 96.09 $\pm$ 1.41 (+2.86)     \\ \hline \hline
Fusion - Naïve~\cite{tacsl2024}       &            & \checkmark       &                & \checkmark     & \checkmark       & 92.97 $\pm$ 1.41            \\
Fusion - Gated ($\lambda_{\textrm{sym}}=0$)        &            & \checkmark       &                & \checkmark     & \checkmark       & 94.53 $\pm$ 2.73 (+1.56)     \\
Fusion - CMT ($\lambda_{\textrm{sym}}=0$)          &            & \checkmark       &                & \checkmark     & \checkmark       & 96.22 $\pm$ 0.98 ({\bf{+3.25}})     \\ \hline \hline
Fusion - Gated + Symmetry regularization ($\lambda_{\textrm{sym}}=1$)      &            & \checkmark       &                & \checkmark     & \checkmark       & 95.05 $\pm 1.76$ (+2.08)    \\
Fusion - CMT + Symmetry regularization ($\lambda_{\textrm{sym}}=1$)         &            & \checkmark       &                & \checkmark     & \checkmark       & 96.59 $\pm 2.11$ ({\bf{+3.62}})  \\ \hline
\end{tabular}
\label{tb:perf}
\end{table*}

\subsection{Physics-Informed Symmetry Regularization}
Inspired by biological motor control principles~\cite{morasso1981arm,bizzi1991modular}, we introduce a physics-informed auxiliary loss that enforces bilateral force balance (Fig.~\ref{fig:sym_loss}). 
Specifically, the right tactile map is vertically flipped and encoded as $\tilde{h}_t^R$, which is compared against $h_t^L$ as
\begin{equation}
\mathcal{L}_{\text{sym}} = \mathbb{E}_{t \sim \mathcal{D}} \big[ \| h_t^L - \tilde{h}_t^R \|_2^2 \big].
\end{equation}
This regularization serves two functions: 
(i) \textit{pre-insertion}, it suppresses asymmetric grasp forces and stabilizes initial contact; 
(ii) \textit{during insertion}, it reduces lateral misalignments that otherwise cause jamming.
We therefore interpret $\mathcal{L}_{\text{sym}}$ as a \textbf{physics-informed inductive bias}, injecting physical consistency into the learned policy. This stabilizes grasp forces and mitigates lateral misalignments, while generalizing from symmetric to asymmetric manipulation tasks.

\subsection{Policy Optimization with PPO}
The overall training objective is
\begin{equation}
\mathcal{L} = \mathcal{L}_{\text{PPO}} + \lambda_{\text{sym}} \mathcal{L}_{\text{sym}},
\end{equation}
where $\mathcal{L}_{\text{PPO}}$ is the clipped surrogate objective of PPO, and $\lambda_{\text{sym}}$ balances task reward with regularization.
The stochastic policy outputs Gaussian actions with bounded variance
\begin{equation}
\sigma_\theta(o_t) \leftarrow \text{clamp}(\sigma_\theta(o_t), \sigma_{\min}, \sigma_{\max}),
\end{equation}
ensuring training stability across random seeds. 

Our framework unifies three innovations: (i) intra-modal tactile self-attention for symmetry-aware encoding, (ii) vision-guided cross-modal attention for structured fusion, and (iii) physics-informed symmetry regularization for stable and precise insertion. Together, these components advance visuo-tactile fusion beyond naïve concatenation or gated fusion, achieving near-privileged performance while retaining robustness in realistic, contact-rich scenarios.

\begin{figure*}[t]
  \begin{center}
    \includegraphics[width=6.2in]{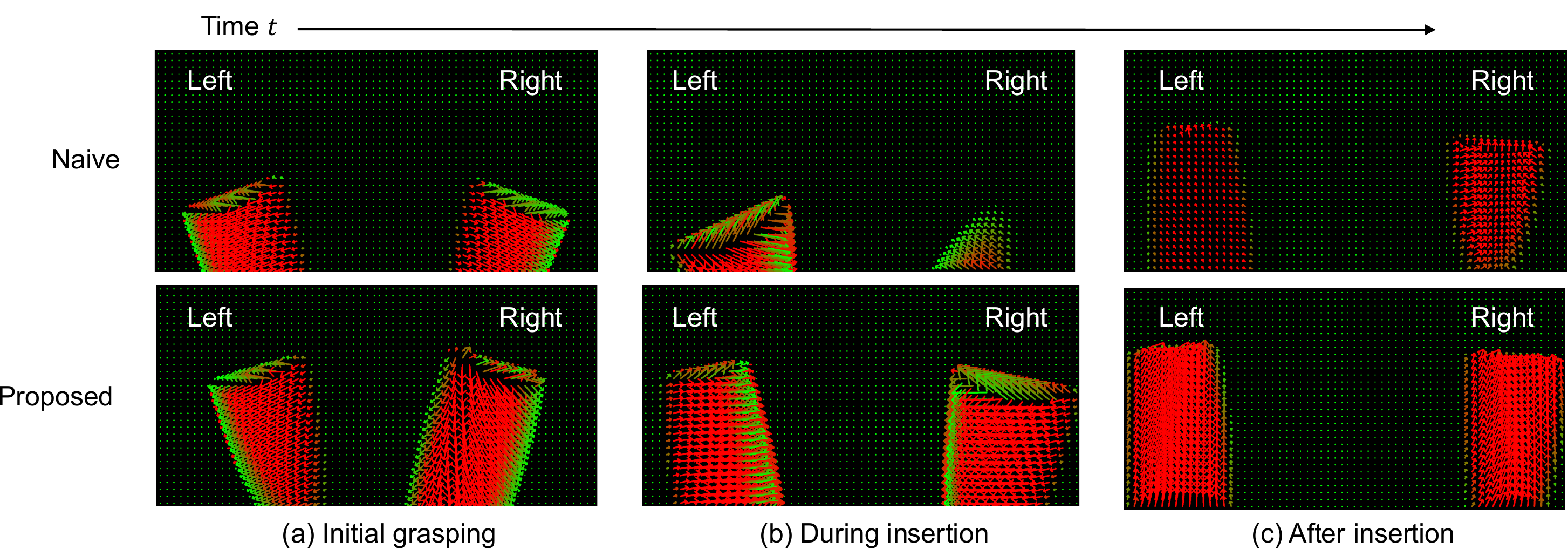}
  \end{center}
  \vspace{-4mm}
    \caption{Evolution of bilateral force fields during insertion under two fusion strategies. \textbf{Top}: Naïve fusion does not enforce symmetry; contact with the socket induces pronounced left–right imbalance, triggering unstable corrections and occasional re-grasping. \textbf{Bottom}: The proposed symmetry-aware CMT maintains balanced force distributions throughout the episode, reducing unnecessary lateral contact and yielding a straighter, smoother insertion trajectory aligned with the table normal. This illustrates how explicitly modeling bilateral symmetry stabilizes contact-rich manipulation under visuo-tactile fusion.}
  \label{fig:qual}
  \vspace{-4mm}
\end{figure*}

\section{Experiments}
We evaluate the proposed symmetry-aware visuo-tactile fusion framework on the robotic insertion task. We first describe the experimental setup, including environment and sensors, followed by the baselines considered for comparison. We then present quantitative results on insertion success rates and qualitative analyses highlighting the role of tactile sensing and symmetry priors. All experiments are conducted with three different random seeds, and reported values correspond to averages across these runs with standard deviations.

Following the evaluation framework established in TacSL~\cite{tacsl2024}, we adopt the same insertion task setup and success criteria (cf. their Table~V, Table~VI, and Table~VII). This ensures consistency in benchmarking and enables a direct comparison of our method against prior visuo-tactile fusion approaches. A detailed description of the experimental setup, sensor configurations, and additional results is provided in the Appendix~\ref{sec:appendix1}.
To foster reproducibility, all training code and task configurations will be released publicly in the official IsaacGymEnvs repository\footnote{\url{https://github.com/isaac-sim/IsaacGymEnvs}}.

\subsection{Quantitative Analysis} \label{sec:quant}
Table~\ref{tb:perf} reports insertion success rates across different observation modalities and fusion strategies. A first observation is that incorporating contact force sensing consistently improves performance. In the privileged setting with compact state features, adding contact forces increases success from $96.74\%$ to $98.96\%$, a $+2.22\%$ absolute gain. A similar trend is observed in the vision-based policy: augmenting the wrist camera with contact force raises success from $93.23\%$ to $96.09\%$ ($+2.86\%$). These results indicate that contact feedback conveys information—such as micro-slippage and compliance—that vision alone cannot provide.

Tactile sensing also demonstrates strong standalone performance: the tactile-only policy reaches 91.41\%, showing that rich geometric and force-distribution information is directly embedded in tactile signals, even without visual context. This robustness suggests that tactile feedback can sustain task execution under degraded or occluded visual conditions.

The most significant improvement arises in visuo-tactile fusion. While naïve concatenation and gated fusion yield only moderate gains over unimodal policies, the proposed CMT achieves $96.22\%$, substantially surpassing both baselines. Notably, this performance nearly matches the privileged ``wrist + contact force'' configuration ($96.09\%$), underscoring that structured cross-modal attention can extract benefits previously attainable only with privileged supervision.
Furthermore, adding the proposed symmetry regularization boosts performance in both gated and CMT architectures. For gated fusion, the success rate increases from $94.53\%$ to $95.05\%$, while for CMT it rises from $96.22\%$ to $96.59\%$. These gains, though modest, indicate that symmetry-aware balancing acts as a stabilizing inductive bias, reducing variability across seeds and further narrowing the gap to privileged sensing.

In summary, three insights emerge: (i) contact forces are indispensable for precise insertion across all policy types; (ii) tactile sensing is independently informative and offers resilience when vision is limited; and (iii) principled fusion via CMT unlocks the full potential of multimodal perception, narrowing the gap to privileged force feedback.

\subsection{Qualitative Analysis}\label{sec:qual}
Fig.~\ref{fig:qual} contrasts the evolution of left/right tactile force fields for naïve versus symmetry-aware visuo-tactile fusion. In the naïve case (top row), contact events with the socket wall disturb the balance between the two fingers; the resulting lateral forces and torques manifest as asymmetric flow patterns over the tactile arrays, often followed by corrective re-grasping and attitude oscillations. These behaviors are typical when fusion does not exploit structure, echoing prior observations that simple concatenation frequently fails to capitalize on multimodal complementarity in contact-rich tasks \cite{tacsl2024}.

In contrast, the proposed CMT-based fusion (bottom row) couples a symmetry-aware tactile encoder with cross-modal attention. By vertically mirroring the right tactile map and aligning it to the left in feature space, the model learns bilateral consistency even under transient contacts; cross-attention then gates these tactile cues with visual context so that pose corrections are made before sustained contact occurs. The resulting force fields remain balanced as the gripper transitions from pre-insertion to insertion, yielding trajectories that are close to straight-down motion, with near-zero net lateral torque—akin to human practice of pre-aligning the plug and then inserting cleanly to minimize incidental contact.

\begin{figure}[t]
  \begin{center}
    \includegraphics[width=3.2in]{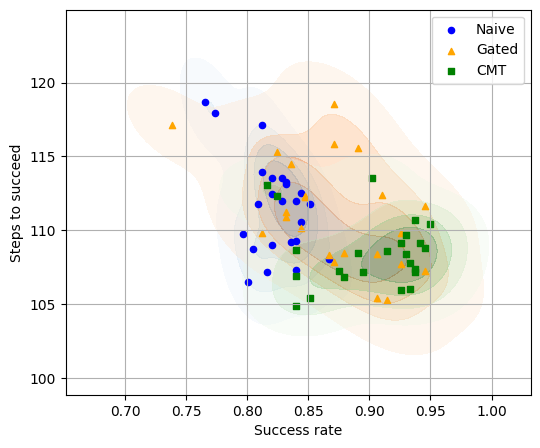}
  \end{center}
  \vspace{-4mm}
    \caption{Distributions of insertion performance for Naïve (blue), Gated (orange), and CMT (green). Scatter points denote individual trials, with kernel density contours indicating outcome distributions in terms of success rate (x-axis) and steps to succeed (y-axis).}
  \label{fig:steps}
  \vspace{-4mm}
\end{figure}

To quantify these qualitative observations, Fig.~\ref{fig:steps} reports the distribution of steps required to complete insertion. The naïve fusion policy requires $111.63$ steps on average, whereas CMT reduces this to $108.48$ steps ($-2.83\%$). Importantly, this reduction directly follows from the balanced force distributions observed in Fig.~\ref{fig:qual} by maintaining symmetry throughout insertion, CMT reduces corrective oscillations, which shortens the trajectory and yields more efficient execution. This provides a clear causal link between qualitative stability and quantitative improvements.


\begin{table}[t]
\footnotesize \centering
\caption{Inference latency, memory usage, and throughput of different methods. Latency is measured over 1,000 forward passes after warm-up, and throughput is computed as $1000 / \textrm{latency}$.}
\vspace{-1mm}
\begin{tabular}{lccc} \hline
Method & Latency (ms) & Memory (MB) & Throughput (fps) \\ \hline
Naïve  & 5.42          & 19.24      & 184.50               \\
Gated  & 5.51          & 17.43      & 181.49               \\
CMT    & 6.52          & 21.45      & 153.37               \\ \hline
\end{tabular}
\label{tb:speed}
\vspace{-4mm}
\end{table}

\subsection{Computation Analysis}\label{sec:perf}
Finally, we assess the computational efficiency of the fusion models. Table~\ref{tb:speed} reports latency, throughput, and memory usage, averaged over $1{,}000$ forward passes. Throughput is computed as $1000 / \textrm{latency (ms)}$.

The naïve model achieves the lowest latency (5.42 ms, 184.5 fps). The gated model shows comparable runtime (5.51 ms, 181.5 fps) while achieving the smallest memory footprint (17.43 MB). The proposed CMT, which incorporates cross-modal Transformer layers, incurs slightly higher cost (6.52 ms, 153 fps, 21.45 MB), yet remains well within real-time control requirements of $60$ Hz.

Although CMT incurs $\sim$20\% higher latency than the naïve baseline, this trade-off is justified by its substantial $+3.62\%$ improvement in success rate (Table~\ref{tb:perf}). Importantly, all models exceed 150 fps, indicating that performance differences are not constrained by computation but rather by the ability to exploit multimodal structure effectively.

Overall, naïve and gated fusion offer minimal overhead, but the proposed CMT provides the best trade-off between computational efficiency and performance, making it the most practical choice for real-world deployment.

\section{Conclusion}
We proposed a visuo-tactile fusion framework for robotic insertion that combines cross-modal attention with a physics-informed balancing loss. Experiments demonstrate three consistent insights: (i) tactile sensing is indispensable for precise alignment, (ii) tactile-only policies remain robust under degraded vision, and (iii) structured fusion via a Cross-Modal Transformer achieves near-privileged performance in real time. These findings establish tactile sensing as a \textbf{core modality for contact-rich manipulation}.

While validated here on symmetric plug insertion, the formulation generalizes naturally to asymmetric objects through reference calibration and data-driven invariances. Future work will extend evaluation to such tasks. By uniting the indispensability of tactile feedback with the stability of physics-informed regularization, this work lays the foundation for \textbf{general visuo-tactile policies} that adapt across diverse geometries and real-world assembly scenarios.

To foster reproducibility, all code and configuration files will be released, enabling direct replication of our results.
\section*{APPENDIX}
\section{Training Details and Experimental Setup}\label{sec:appendix1}

We follow the training setup and experimental design of TacSL~\cite{tacsl2024}, and reproduce the key details here for completeness. This ensures that our results are directly comparable to their benchmarks, and that the proposed visuo-tactile fusion framework can be reproduced independently of their paper.

\begin{table}[t]
\centering
\caption{Environment randomization bounds (adapted from TacSL Appendix C, Table V). Each parameter is uniformly sampled within the specified range.}
\label{tab:rand_bounds}
\begin{tabular}{ll}
\hline
Parameter & Range \\ \hline
End-effector X (m) & [0.4, 0.6] \\
End-effector Y (m) & [-0.1, 0.1] \\
End-effector Z (m) & [0.1, 0.2] \\
End-effector Euler-X (rad) & [3.04, 3.24] \\
End-effector Euler-Y (rad) & [-0.1, 0.1] \\
End-effector Euler-Z (rad) & [-1.0, 1.0] \\
Socket X (m) & [0.4, 0.6] \\
Socket Y (m) & [-0.1, 0.1] \\
Socket Z (m) & [0.0, 0.02] \\
Peg-in-gripper Z-pos (m) & [-0.0125, 0.0125] \\
Peg-in-gripper X-rot (rad) & [-0.628, 0.628] \\
Socket XYZ noise (m) & [-0.005, 0.005] \\
Compliance stiffness noise (N/m) & [150, 350] \\
Compliance damping noise (N/(m/s)) & [0.0, 1.0] \\
Joint damping noise (N/(m/s)) & [-1.5, 1.5] \\ \hline
\end{tabular}
\end{table}

\begin{table}[t]
\centering
\caption{Policy architecture. Vision input is $64\times64\times3$, tactile input is $32\times32\times3$ (3 channels for $f_x,f_y,f_z$). 
All modalities share the same CNN encoder structure; output is projected to a 128-d embedding. Fusion module differs by method.}
\label{tab:policy_backbone}
\resizebox{\linewidth}{!}{
\begin{tabular}{l|c|c}
\hline
Layer / Module & Configuration & Output Dim. \\ \hline
\multicolumn{3}{c}{Encoder (per modality)} \\ \hline
Input (Vision) & $64\times64\times3$ & $(B,3,64,64)$ \\
Conv1 & $8\times8$, stride 2, 32 channels & $(B,32,29,29)$ \\
Conv2 & $4\times4$, stride 1, 64 channels & $(B,64,26,26)$ \\
Conv3 & $3\times3$, stride 1, 64 channels & $(B,64,24,24)$ \\
Spatial SoftArgMax & $64$ channels $\times$ 2D coords & $(B,128)$ \\ \hline
Input (Tactile) & $32\times32\times3$ & $(B,3,32,32)$ \\
Conv1 & $8\times8$, stride 2, 32 channels & $(B,32,13,13)$ \\
Conv2 & $4\times4$, stride 1, 64 channels & $(B,64,10,10)$ \\
Conv3 & $3\times3$, stride 1, 64 channels & $(B,64,8,8)$ \\
Spatial SoftArgMax & $64$ channels $\times$ 2D coords & $(B,128)$ \\ \hline
\multicolumn{3}{c}{Fusion (varies by method)} \\ \hline
Naïve Fusion & Concatenate [vision, left, right] & $(B,384)$ \\
Gated Fusion & Weighted sum [vision, left, right] & $(B,128)$ \\
CMT Fusion & Self-attn + cross-attn & $(B,256)$ \\ \hline
\multicolumn{3}{c}{Policy Head} \\ \hline
RNN & 2-layer LSTM, 256 hidden units & $(B,256)$ \\
MLP & [256, 128, 64] + ELU & $(B,64)$ \\
Policy Mean & Linear $\to \mathbb{R}^6$ & $(B,6)$ \\
Policy Log-Std & Linear $\to \mathbb{R}^6$ & $(B,6)$ \\
Value Head & Linear $\to \mathbb{R}^1$ & $(B,1)$ \\ \hline
\end{tabular}}
\end{table}

\subsection{Environment Randomization}
Table~\ref{tab:rand_bounds} summarizes the task randomization levels applied when generating initial states for the insertion task.  
The end-effector is randomized around a nominal home pose, the peg is initialized with random offsets inside the gripper, and the socket is placed with randomized position in front of the robot.  
Additionally, contact parameters and damping coefficients are randomized, and observation noise is applied to socket localization to simulate imperfect perception.

\subsection{Policy Model Architecture}
Following TacSL~\cite{tacsl2024}, we use a lightweight 3-layer CNN encoder with a Spatial SoftArgMax layer. 
The CNN extracts local features from visual or tactile inputs, while the SoftArgMax produces differentiable keypoint-like embeddings, preserving geometric information critical for insertion. 
On top of this shared encoder, we implement three fusion strategies (naïve, gated, and CMT), combined with a recurrent and feedforward policy head. 
The complete architecture is summarized in Table~\ref{tab:policy_backbone}.

\begin{table}[t]
\centering
\caption{A2C hyperparameters.} \label{tab:ppo_hyp}
\begin{tabular}{ll}
\hline
Parameter & Value \\ \hline
Optimizer & Adam \\
Learning rate & 1.0e-4 \\
Rollout / Horizon length & 512 steps \\
Mini-batch size & 512 \\
PPO / Mini epochs & 4 \\
Discount factor $\gamma$ & 0.99 \\
GAE parameter $\lambda$ & 0.95 \\
Clip ratio $\epsilon$ & 0.2 \\
Max gradient norm & 1.0 \\
Entropy coefficient & 0.0 \\
Bounds loss coefficient & 0.0001 \\
Value loss / Critic coefficient & 2 \\ \hline
\end{tabular}
\vspace{-4mm}
\end{table}

\subsection{Training Hyperparameters}
We train all policies with PPO using the hyperparameters listed in Table~\ref{tab:ppo_hyp}. 
These are directly taken from TacSL~\cite{tacsl2024} Appendix C (Table VII).

\subsection{Code and Configuration Availability}
For completeness, we follow the official TacSL task configuration provided in their public repository.  
The full environment specification for the insertion task (including randomization, observation spaces, and reward shaping) is available in the TacSL branch of the IsaacGymEnvs repository at TacSLTaskInsertion.yaml\footnote{\url{https://github.com/isaac-sim/IsaacGymEnvs/blob/tacsl/isaacgymenvs/cfg/task/TacSLTaskInsertion.yaml}}.

\section{Tactile Sensing for Screw Tasks}\label{sec:appendix2}

To further assess the role of tactile sensing, we extend our evaluation to the screw task, which involves grasping a nut and performing screwing motions onto a bolt. We follow the experimental setup and environment specifications described in Factory~\cite{chi2022factory}, ensuring consistency with prior benchmarks.

\begin{figure}[t]
  \begin{center}
    \includegraphics[width=3.2in]{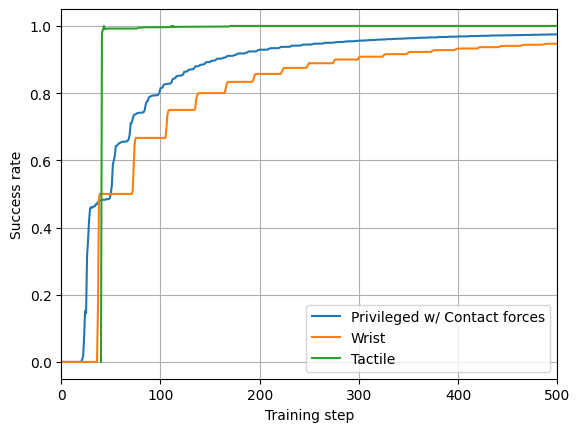}
  \end{center}
  \vspace{-4mm}
    \caption{Insertion performance for the screw task under three policies: Privileged (blue), Wrist-camera (orange), and Tactile (green). The high-resolution tactile policy achieves perfect success, outperforming even the privileged contact-force baseline, demonstrating the benefit of detailed tactile feedback for precise contact-rich manipulation.}
  \label{fig:screw}
  \vspace{-4mm}
\end{figure}

Our results, summarized in Fig.~\ref{fig:screw}, reveal that tactile sensing alone is sufficient to solve the screw task with perfect reliability. Specifically, the tactile-only policy achieves a $100\%$ success rate across all evaluated trials, outperforming both the privileged contact-force baseline and the wrist-camera policy. This advantage likely stems from the tactile arrays providing a $3\times32\times32$ measurement of surface deformations, capturing rich geometric and force details. In contrast, the privileged baseline only observes a compact $3\times1$ contact-force vector.

These findings suggest that for manipulation tasks dominated by fine-grained contact interactions, high-resolution tactile feedback can serve as a complete substitute for vision. While vision may still be beneficial for broader alignment or multi-step tasks, Fig.~\ref{fig:screw} demonstrates that tactile sensing alone is sufficient for precise screw insertion, highlighting its potential as a primary modality for contact-intensive operations.

Expanding the task scope to include both \emph{insert} (lifting and aligning the nut onto the bolt) and \emph{screw} (rotational insertion of the nut) stages will likely restore the necessity of multimodal fusion. In such settings, vision can provide global pose and alignment context, while tactile sensing ensures precise contact tracking during screwing. We expect that principled visuo-tactile fusion will thus be critical for scaling toward real-world robot assembly, where robustness, generalization, and efficiency are essential for industrial deployment.

\end{document}